\documentclass[letterpaper]{article} 
\usepackage[preprint]{aaai2027}  
\usepackage[hyphens]{url}  
\usepackage{graphicx} 
\usepackage{natbib}  
\usepackage{caption} 
\usepackage{booktabs}
\usepackage{multirow}
\usepackage{pifont}
\usepackage{array}
\usepackage{makecell}
\usepackage{tabularx}
\usepackage{xcolor}
\usepackage{amsmath}
\usepackage{amssymb}
\usepackage{amsfonts}
\usepackage{subcaption}
\usepackage{xurl}
\usepackage{threeparttable}
\usepackage{amsthm}
\newtheorem{definition}{Definition}

\DeclareCaptionStyle{ruled}{labelfont=normalfont,labelsep=colon,strut=off} 
\title{Content Depth Matters in Short-Video Recommendation: \\ Rethinking the Attention Economy}

\author{
    Liwei Deng\textsuperscript{\rm 1},
    Jing Jiang\textsuperscript{\rm 1},
    Zhiwei Li\textsuperscript{\rm 1}
    \protect\thanks{Corresponding author: Zhiwei Li
    (zhw.li@outlook.com).},
    Yang Wang\textsuperscript{\rm 2},
    Guodong Long\textsuperscript{\rm 1}
}

\affiliations{
    \textsuperscript{\rm 1}
    Australian Artificial Intelligence Institute,
    University of Technology Sydney\\
    \textsuperscript{\rm 2}
    Evidence and Research,
    Department of Health, Disability and Ageing\\
    liweidengdavid@gmail.com,
    jing.jiang@uts.edu.au,
    zhw.li@outlook.com,
    Alvin.Wang@health.gov.au,
    guodong.long@uts.edu.au,
}

\begin{document}

\maketitle

\begin{abstract}
Driven by the attention economy, short-video Recommender Systems (RSs) are primarily optimized to maximize user engagement by promoting videos that capture attention within seconds. These systems inherently favor shallow-content videos that are effective at attracting immediate attention. However, growing evidence suggests that prolonged exposure to such content may negatively affect users' cognitive engagement and mental well-being, raising concerns about the long-term societal impact of the short-video platform.
To tackle this challenge, this paper introduces a new metric, the \textbf{Content Depth Score (CDS)}, to quantify the content depth of short videos. CDS measures the extent to which a video is expected to stimulate higher-order cognitive processes, using a seven-level scale grounded in established theories of cognitive psychology and learning. As an initial step toward this vision, we present \textbf{SCOPE-Bench}, the first benchmark for content-depth evaluation in short-video recommendation. Built upon a large-scale open-source short-video dataset, SCOPE-Bench provides CDS annotations for 150K videos, enabling systematic evaluation of RSs from a cognitive-content perspective.
Leveraging SCOPE-Bench, we evaluate 13 representative RSs and reveal a consistent preference for shallow-content videos. Moreover, we find that these algorithms recommending cognitively deep content are only marginally better than random selection, highlighting a previously overlooked limitation of existing recommendation objectives.
Our code and datasets are available at \url{https://liweidengdavid.github.io/SCOPE-Bench/}.

\end{abstract}

\section{Introduction}
In the attention economy, short-video feeds on platforms such as YouTube \cite{YouTube}, TikTok \cite{TikTok}, and Kuaishou \cite{Kuaishou} have become a central channel for everyday content consumption.
Adult users now spend more than one hour per day on these platforms \cite{short_video_time}.
Their Recommender Systems (RSs) \cite{RS_Survey2,FedVLR} optimize engagement signals, such as watch time and clicks, to turn user attention into revenue \cite{Attention_Economy}, and therefore favor videos capable of rapidly capturing users’ attention.
Over time, continuous exposure to such content may weaken users' sustained attention \cite{Attention_Pattern} and affect their long-term well-being \cite{Mental_health}.

\begin{figure}[t]
    \centering
    \includegraphics[width=1\linewidth]{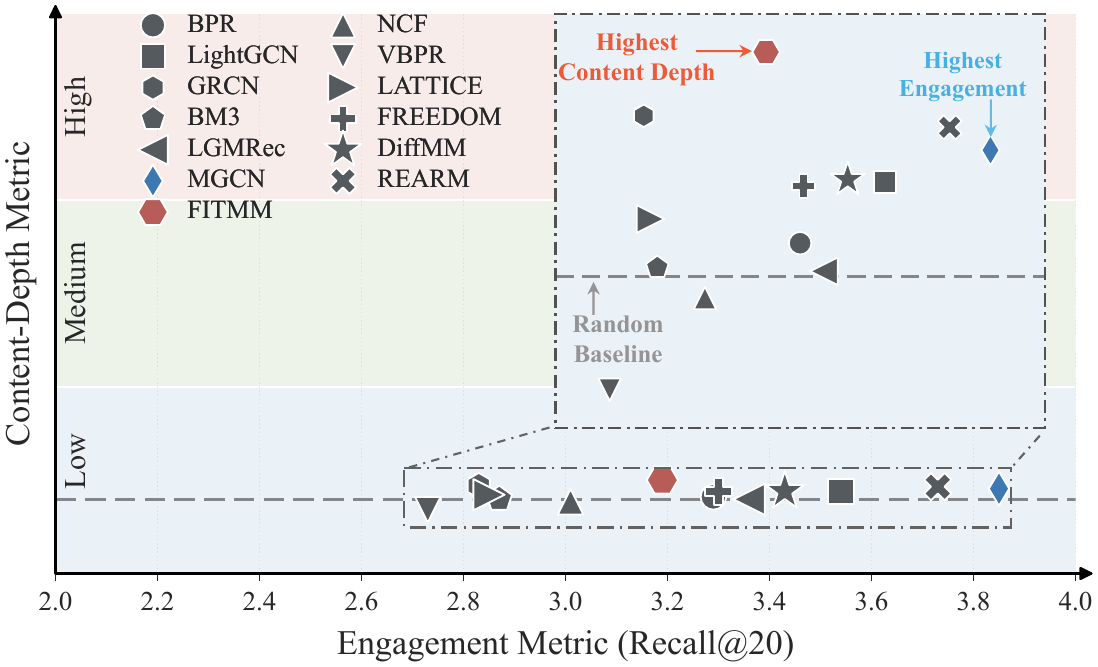}
    \caption{Evaluating 13 Recommender Systems (RSs) on both
    Engagement metric (X axis) and our Content-Depth metric (Y axis).
    The regions are split into three levels corresponding to content-depth metric: Low (blue), Medium (green), and High (red). The dashed line marks the content depth of a random selection baseline.
    Existing RSs perform competitively on the engagement metric while their content-depth metric is low and near to the random selection.
    }
    \label{fig:intro}
\end{figure}

Optimizing for engagement rewards videos that retain user attention, which raises a more fundamental question:
\textit{Do RSs favor attention-grabbing content over videos with greater depth?}
Content depth reflects how deeply a video develops its information, providing users with richer opportunities for understanding, reasoning, and reflection \cite{ICAP}.
We examine this question by comparing the engagement of 13 representative RSs  with the content depth of the videos they recommend.
As Figure~\ref{fig:intro} shows, the content-depth performance of every RS remains close to that of random recommendation.
Therefore, stronger engagement does not translate into greater recommended content depth.

The situation has already drawn awareness and responses beyond research.
For example, Australia~\cite{AustraliaSocialMediaMinimumAge2025} and the UK~\cite{UKChildrenOnlineRules2026} have imposed under-16 restrictions on short-video platforms, and platforms themselves cap adolescents' daily usage \cite{Usage_time}.
This convergence of regulators signals that the harm to young users is now widely acknowledged.
Yet these interventions act on access rather than content.
One important reason is the absence of any measure for the content itself.

Inspired by the need to promote healthier short video and build a sustainable short-video ecosystem, we propose a new metric to make content depth measurable, enabling it to serve as an explicit objective in RS evaluation and optimization.
This paper proposes the \textbf{C}ontent \textbf{D}epth \textbf{S}core (\textbf{CDS}), a novel metric to measure how deeply a short video presents and delivers information to humans.
CDS scores a video on a seven-level rubric grounded in theories of cognition and learning~\cite{Dual_process, Bloom_Taxonomy, SOLO_Taxonomy}, ranging from low-level emotional stimulation to higher-order cognitive processes\footnote{
Specifically, in this paper, cognitive processes refer to the mental operations through which individuals acquire, process, store, and use information \cite{Cognitive}.
CDS measures it as the opportunities a video's content provides for these operations.}.
To validate CDS at both the item and recommendation-list levels, we construct \textbf{SCOPE-Bench}, a \textbf{S}hort-video \textbf{CO}ntent de\textbf{P}th \textbf{E}valuation \textbf{Bench}mark, by extending the existing open-source ShortVideo dataset \cite{ShortVideo}.
Specifically, we annotate its 150K videos with CDS labels, covering about 1M user-item interactions from 10K users.

Our main contributions are summarized as follows:
\begin{itemize}
\item
To the best of our knowledge, we are the first to formulate and systematically investigate the lack of attention to content depth in engagement-optimized RSs.

\item
We propose CDS, the first quantitative metric for video content depth, based on a seven-level rubric grounded in theories of cognition and learning.
We further extended CDS to a list-level metric for recommended lists.

\item
We establish a comprehensive evaluation framework for video content depth and construct SCOPE-Bench, a benchmark of 150K videos with human-aligned CDS annotations and about 1M interactions from 10K users.

\item
Experiments on SCOPE-Bench confirm that the CDS evaluation protocol closely aligns with human judgments.
Using CDS to evaluate 13 representative RSs, we find that their recommended lists achieve content-depth performance close to that of random recommendation, revealing that engagement and content depth are decoupled.

\end{itemize}

\section{Related Works}

\subsection{Content Quality Assessment}
Content quality assessment can be broadly categorized into \emph{metric-based assessment} for explicitly quantifiable properties and \emph{judgment-based assessment} for open-ended or interpretive properties that are difficult to capture with conventional metrics.
The former mainly covers \textbf{video quality assessment}, including perceptual fidelity~\cite{SSIM}, temporal consistency~\cite{VideoQualityStructuralDistortion}, generative realism~\cite{FID}, and cross-modal alignment~\cite{CLIPSIM}, as well as \textbf{text quality assessment}, covering linguistic~\cite{Fluency} and semantic properties~\cite{Coherence}, trustworthiness~\cite{Factuality}, and safety-related dimensions~\cite{Toxicity}.
For open-ended and interpretive properties, recent studies have increasingly adopted \textbf{LLM-as-a-Judge} as a flexible judgment-based evaluation paradigm.
Existing approaches range from scalar scoring \cite{G-eval} and pairwise comparison~\cite{LLM-as-a-judge} to rubric-based evaluation~\cite{Prometheus,Pandalm} and fine-grained assessment protocols~\cite{Flask}.
While effective, these methods provide only a limited conceptualization of content depth.
In the paper, we systematically define content depth and propose the first quantitative metric, termed the CDS, to measure the content depth.

\subsection{Evaluation of Recommender Systems}
Existing evaluation of RSs can be broadly categorized into \emph{system-centric evaluation} and \emph{user-centric evaluation}.
System-centric evaluation covers \textbf{accuracy-oriented evaluation} and \textbf{beyond-accuracy evaluation} \cite{Evaluating_RS_Survey}.
The former assesses whether relevant items are accurately retrieved and ranked using metrics such as Precision~\cite{Precision}, Recall~\cite{Recall}, and NDCG~\cite{NDCG}, whereas the latter considers complementary recommendation qualities, including diversity~\cite{diversiy}, novelty~\cite{novelty}, and catalog coverage~\cite{catalog_coverage}.
In contrast, \textbf{user-centric evaluation} examines users' subjective perceptions and experiences, including choice difficulty~\cite{choice_difficulty}, usefulness~\cite{usefulness}, and satisfaction~\cite{satisfaction}.
Although prior studies have attempted to evaluate RSs beyond conventional engagement metrics, little research has quantitatively examined the content depth of recommended videos.
To address this gap, we introduce the List-wise CDS (LCDS), the first list-level metric for quantifying the content depth of recommended lists.

\section{Content Depth Score Metric}
\begin{table*}[t]
\centering
{
\footnotesize
\renewcommand{\arraystretch}{1.1}
\setlength{\tabcolsep}{1.mm}
\renewcommand{\tabularxcolumn}[1]{m{#1}}
\setlength{\belowrulesep}{0.25ex}
\begin{tabularx}{\textwidth}{
    @{}
    >{\centering\arraybackslash}m{1.12cm}
    >{\centering\arraybackslash}m{1.75cm}
    >{\raggedright\arraybackslash}X
    >{\centering\arraybackslash}m{1.65cm}
    >{\centering\arraybackslash}m{2.25cm}
    >{\centering\arraybackslash}m{2.65cm}
    @{}
}

\specialrule{\heavyrulewidth}{0pt}{-1.1ex}

\makecell[c]{\textbf{CDS}\\\textbf{Level}}
&
\makecell[c]{\textbf{Rubric}\\\textbf{Label}}
&
\makecell[l]{\textbf{Operational}\\\textbf{Criterion}}
&
\makecell[c]{\textbf{Dual} \textbf{Process}\\ \textbf{Theory}}
&
\makecell[c]{\textbf{Bloom's}\\\textbf{Taxonomy}}
&
\makecell[c]{\textbf{SOLO}\\\textbf{Taxonomy}}
\\

\midrule

\textbf{Level 0}
&
\textbf{Affect}
&
Mainly evokes affect, humor, spectacle, or atmosphere.
&
System 1
&
--
&
--
\\

\addlinespace[0.15em]

\textbf{Level 1}
&
\textbf{Point}
&
Presents an isolated opinion, or label without explaining why or how.
&
System 2
&
Remember
&
Prestructural
\\

\textbf{Level 2}
&
\textbf{Concept}
&
Defines or illustrates a single idea with simple background or explanation.
&
System 2
&
Understand
&
Unistructural
\\

\textbf{Level 3}
&
\textbf{Procedure}
&
Shows how a method can be used in concrete cases, typically through multiple steps or illustrative examples.
&
System 2
&
Apply
&
Multistructural
\\

\addlinespace[0.15em]

\textbf{Level 4}
&
\textbf{Mechanism}
&
Explains mechanisms, variables, constraints, conditions, causal links, or system relationships.
&
System 2
&
Analyze
&
Relational
\\

\textbf{Level 5}
&
\textbf{Judgment}
&
Weighs evidence or competing explanations, including limitations, uncertainty, or counterexamples.
&
System 2
&
Evaluate
&
\makecell[c]{Relational /\\Extended Abstract}
\\

\textbf{Level 6}
&
\textbf{Model}
&
Builds a generalizable model, framework, principle, or decision rule that can transfer across contexts.
&
System 2
&
Create
&
Extended Abstract
\\

\bottomrule

\end{tabularx}
}

\caption{
Seven-level scoring rubric for CDS and its approximate theoretical anchors.
Each level is defined by an operational criterion for consistent annotation.
A dash indicates that no direct theoretical correspondence is assigned.
The rubric captures a progression from affective responses to increasingly complex reasoning and transferable model construction.
}
\label{tab:cognitive_rubric}

\end{table*}

\subsection{Content Depth}

To measure content depth, we begin by defining it precisely.
\begin{definition}[Content Depth]
\label{def:content_depth}
Content depth of a video is the degree to which it develops a topic from isolated information into structured understanding by explaining concepts, demonstrating procedures, analyzing mechanisms, forming evaluative judgments and generalizable insights.

\end{definition}
Based on Definition~\ref{def:content_depth}, we propose CDS, the first metric to quantify the content depth of short videos.
A video receives a higher CDS when it presents richer semantic information, clearer explanations, and higher-order reasoning structures.
Content depth is cognitively meaningful because deeper content offers viewers richer opportunities for higher-order cognitive processes~\cite{Bloom_Taxonomy},
which support cognitive development and intellectual growth.

\subsection{Seven-level Scoring Rubric}

\textbf{Theoretical Foundation.}
We operationalize CDS as a seven-level rubric grounded in Dual Process Theory~\cite{Dual_process}, the Revised Bloom's Taxonomy\footnote{For simplicity, we refer to it as ``Bloom's Taxonomy'' hereafter.}\cite{Bloom_Taxonomy}, and the SOLO Taxonomy~\cite{SOLO_Taxonomy}.
Specifically, Dual Process Theory distinguishes between immediate affective and deliberate processing.
Bloom's Taxonomy organizes cognitive processes into six levels of increasing complexity, whereas the SOLO Taxonomy classifies the degree of understanding demonstrated by learners into five levels.
Based on these complementary perspectives, we use System~1 in Dual Process Theory to describe the lowest level and the six cognitive levels of Bloom's Taxonomy to define the remaining levels.
We further use the SOLO Taxonomy as complementary guidance for distinguishing the complexity of understanding across levels.
The resulting rubric ranges from immediate affective responses to increasingly complex cognitive processes.

\textbf{Rubric Structure.}
Table~\ref{tab:cognitive_rubric} presents the full rubric and its approximate theoretical anchors, grouping the seven levels into three groups.
Low-CDS group (Level 0)
represents content that mainly elicits affective or entertaining responses, with limited explicit knowledge value.
Medium-CDS group (Levels 1--3)
captures progressively more structured knowledge transmission, moving from isolated information to conceptual explanation and practical application.
High-CDS group (Levels 4--6)
reflects higher-order cognitive processes, ranging from analytical reasoning to evaluative judgment and the development of transferable models or frameworks.

Detailed indicators and examples, together with the theoretical foundations and construction of the rubric, are provided in Appendices~A and~B, respectively.

\section{Benchmark for CDS Evaluation}

\begin{figure*}[t]
    \centering
    \includegraphics[width=0.8\linewidth]{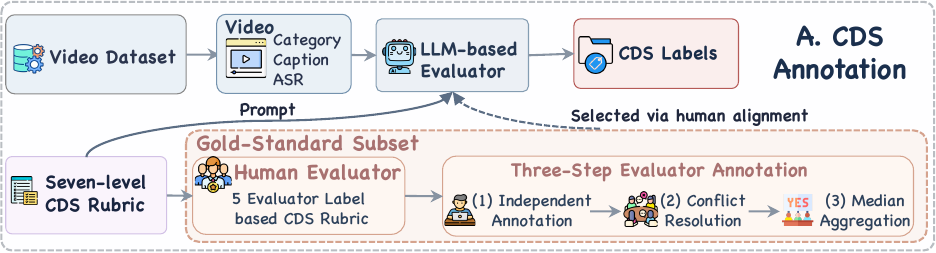}
   \caption{Overview of the scalable CDS annotation workflow, which applies the rubric grounded in theories of cognition and learning.
   A gold-standard subset is used to select the LLM evaluator best aligned with human judgments for large-scale labeling.}
    \label{fig:Overview_a}
\end{figure*}

We introduce SCOPE-Bench, the first benchmark that supports content-depth assessment from two different perspectives.
At the \textbf{item level} (Figure~\ref{fig:Overview_a}), SCOPE-Bench assesses the content depth of individual short videos from the {user perspective}.
At the \textbf{recommendation-list level} (Figure~\ref{fig:Overview_b}), it evaluates the content depth of recommendation lists from the {platform perspective}, enabling platforms to assess RSs beyond conventional engagement-oriented performance.

\begin{table}[tb]
\centering
\setlength{\tabcolsep}{1.5pt}
\renewcommand{\arraystretch}{0.85}
{\fontsize{8.1pt}{10pt}\selectfont
\begin{tabular*}{\linewidth}{
  @{\extracolsep{\fill}}
  *{8}{c}
  @{}
}
\toprule
\textbf{Dataset}
& \textbf{\#Users}
& \textbf{\#Items}
& \textbf{\#Inter.}
& \textbf{Caption}
& \textbf{Image}
& \textbf{ASR}
& \textbf{Sparsity} \\
\midrule

\textsc{Full}
& 10,000
& 153,561
& 1,007,746
& 100.00\%
& 88.56\%
& 88.56\%
& 99.93\% \\

\textsc{Sampled}
& 6,654
& 31,496
& 128,105
& 100.00\%
& 99.26\%
& 99.26\%
& 99.94\% \\

\bottomrule
\end{tabular*}
}

\vspace{2pt}

\begin{minipage}{\linewidth}
{\fontsize{8pt}{10pt}\selectfont
\textit{Note:}
The modality coverage is not complete because captions, visual features,
and Automatic Speech Recognition (ASR) transcripts in the publicly released
data are not uniformly available for all videos.
}
\end{minipage}

\caption{
Statistics and modality coverage of the dataset.
}
\label{tab:dataset_statistics}
\end{table}

\subsection{Video Dataset}
We build SCOPE-Bench upon ShortVideo\footnote{\url{https://github.com/tsinghua-fib-lab/ShortVideo_dataset}}~\cite{ShortVideo}, a publicly available short-video dataset containing user-item interactions and multimodal information.
ShortVideo records approximately 1M chronologically ordered interactions generated by 10K real-world users over a one-week period.
The dataset provides two versions, \textsc{Full} and \textsc{Sampled}, which cover the same data collection period.
The \textsc{Sampled} version contains a subset of the users included in \textsc{Full}, although the sampling strategy used to construct this subset is not documented by the \cite{ShortVideo}.
We perform necessary preprocessing to improve data consistency and retain the content signals required for CDS assessment, as detailed in Appendix~C.
The statistics of the resulting \textsc{Full} and \textsc{Sampled} versions are summarized in Table~\ref{tab:dataset_statistics}.

\subsection{Evaluation Protocol}
Our protocol adopts a content-based setting, where the video content is the primary object of CDS assessment.
Instead of directly processing raw visual and audio streams, we use three textual signals as input, which capture complementary aspects of the video content: the caption $c$ summarizes the main theme, the category label $k$ provides topic-level context, and the ASR transcript $a$ captures the spoken content.
Given a video $x=(c,k,a)$ and the system prompt $\mathcal{P}$, an LLM evaluator $\mathcal{M}$ produces a structured output $y=(s,\ell,r,e,q)$, comprising the CDS $s$, its level name $\ell$, a reason $r$, supporting evidence $e$, and a confidence $q$.
The score $s$ is ordinal, taking values in $\{0,1,\dots,6,\varnothing\}$, where $\varnothing$ marks insufficient information for a reliable assessment.
Because this setting relies on text, videos whose depth resides mainly in visual or auditory content, or that lack a usable transcript, receive $\varnothing$.

\subsection{Annotation}
\textbf{Gold-Standard Subset Construction.}
To establish reliable reference labels for CDS assessment, we construct a gold-standard subset (gold set) through human annotation. Specifically, we conducted a human evaluation with five human evaluators from computer science and psychology.
Following the Delphi method~\cite{Delphi}, the annotation process consists of three steps.
In Step~\ding{172}, the evaluators independently annotated the sampled videos according to scoring rubric based on video content.
In Step~\ding{173}, the human evaluators conducted a second round of annotation for videos showing substantial disagreement across evaluators.
In Step~\ding{174}, we aggregated the annotations using the median to reduce the influence of outlier judgments and improve the reliability of the resulting gold-standard labels.
The gold set serves two purposes: validating interpretability and practical usability of our rubric and providing a human reference for selecting the LLM evaluator used in automatic annotation.
Details about human evaluators, the annotation procedure, and the size of the gold set are provided in Appendix~E.

\textbf{Automatic Scalable Annotation.}
To enable scalable annotation while maintaining alignment with human judgments, we first evaluate a set of candidate LLMs on the gold set.
We compare their CDS predictions with human annotations using multiple agreement and error metrics, and select the model with the strongest overall alignment as the default evaluator.
The evaluation results are reported in Section~\ref{sec:LLMs-Human Agreement}.
We then apply the selected evaluator to the short-video dataset to generate CDS annotations.
By augmenting the original dataset with annotations, we construct SCOPE-Bench dataset, which supports two complementary evaluation settings at both item and list levels for recommendation.

\begin{figure*}[!t]
    \centering
    \includegraphics[width=0.9\linewidth]{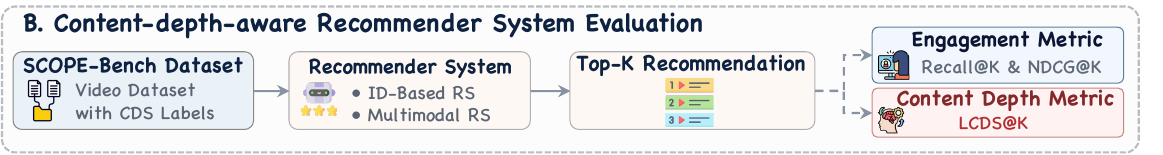}
    \caption{Overview of the content-depth-aware RS evaluation framework for jointly assessing engagement and content depth.}
    \label{fig:Overview_b}
\end{figure*}

\subsection{List-Level Content-Depth Evaluation}
\label{sec:lcpd}
As illustrated in Figure~\ref{fig:Overview_b}, the list-level content-depth evaluation workflow evaluates the content depth of top-$K$ recommendation lists.
Specifically, we introduce \emph{\textbf{L}ist-wise \textbf{C}ontent \textbf{D}epth \textbf{S}core (\textbf{LCDS})}, a complementary metric that aggregates the CDS of the top-$K$ lists.
Let \(s_i \in \{0,\ldots,6\}\)\footnote{We assign a score of 0 to items with $s_i=\varnothing$. Such cases are primarily caused by insufficient information from ASR, so the resulting values should be interpreted as lower-bound estimation.} denote the CDS of the video at rank $i$.
LCDS@$K$ is defined as:
\begin{align}
\operatorname{LCDS}_{\boldsymbol{\alpha},\beta,\boldsymbol{w}}@K
&=\left(\frac{\sum_{i=1}^{K}w_i \left[g_{\boldsymbol{\alpha}}(s_i)\right]^{\beta}}
{\sum_{i=1}^{K} w_i}\right)^{1/\beta}, \label{eq:lcds}\\
\text{where} \;
g_{\boldsymbol{\alpha}}(s)&=
\begin{cases}
0,
& s=0,\\
\displaystyle
\frac{\sum_{\ell=1}^{s}\alpha_\ell} {\sum_{r=1}^{6}\alpha_r}, & s\in\{1,\ldots,6\}.
\end{cases}
\label{eq:normalized_gain}
\end{align}
In Eq.~\eqref{eq:normalized_gain}, $\alpha_\ell>0$ denotes the marginal value of moving from Level $\ell-1$ to Level $\ell$, and $g_{\boldsymbol{\alpha}}(s)\in[0,1]$ maps an ordinal CDS label to a normalized gain.
$\boldsymbol{w}$ model position-dependent exposure and satisfy both $w_i\geq 0$ and $\sum_{i=1}^{K}w_i>0$.

Let $\bar{w}_i={w_i}/{\sum_{j=1}^{K}w_j}$ denote the normalized rank $i$ exposure weight, and $\mathcal{I}_{+} := \left\{ i \in \{1,\ldots,K\} \mid w_i > 0 \right\}$.
The aggregation behavior of LCDS can be characterized by:
\begin{align}
\text{LCDS}_{\beta}@K&
=
\begin{cases}
\displaystyle
\prod_{i\in \mathcal{I}_{+}}
g_{\boldsymbol{\alpha}}(s_i)^{\bar{w}_i},
& \beta \rightarrow 0^{+}, \\[2mm]
\displaystyle
\sum_{i=1}^{K}
\bar{w}_i g_{\boldsymbol{\alpha}}(s_i),
& \beta = 1, \\[2mm]
\displaystyle
\max_{i\in \mathcal{I}_{+}}
g_{\boldsymbol{\alpha}}(s_i),
& \beta \rightarrow \infty.
\end{cases}
\end{align}
Hence, larger values of $\beta$ produce a more peak-oriented evaluation,
whereas smaller values make LCDS more sensitive to low-CDS positions and
therefore favor lists that sustain CDS.
Following the grouping of the original CDS rubric, we further define three corresponding interpretive levels for LCDS.
The thresholds are defined directly on the transformed scale as Low LCDS: $ [0,\tau_{\mathrm{L}})$, Medium LCDS: $[\tau_{\mathrm{L}},\tau_{\mathrm{H}})$, and High LCDS:$[\tau_{\mathrm{H}},1]$, where $\tau_{\mathrm{L}} = g_{\boldsymbol{\alpha}}(1)$, $\tau_{\mathrm{H}} = g_{\boldsymbol{\alpha}}(4)$.

For the default setting, we assume equal marginal values, i.e., \(\boldsymbol{\alpha}=\boldsymbol{1}\), and use arithmetic aggregation with \(\beta=1\), yielding \(g(s_i)=s_i/6\).
Uniform rank weights ($w_i=1$) give Top-$K$ Average LCDS (A-LCDS@$K$) definition as follows:
\begin{equation}
\operatorname{A\text{-}LCDS}@K
=
\frac{1}{K}\sum_{i=1}^{K}g(s_i),
\label{eq:a_lcpd}
\end{equation}
which measures the average content depth of the recommendation list.
Moreover, to account for greater exposure at higher ranks, we set $w_i=1/\log_2(i+1)$ followed by NDCG \cite{NDCG}, and define Top-$K$ Exposure-weighted LCDS (E-LCDS@$K$) as follows:
\begin{equation}
\operatorname{E\text{-}LCDS}@K
=
\frac{
\sum_{i=1}^{K}\frac{g(s_i)}{\log_2(i+1)}
}{
\sum_{j=1}^{K}\frac{1}{\log_2(j+1)}
}.
\label{eq:e_lcpd}
\end{equation}
Both metrics lie in \([0,1]\), with larger values indicating greater content depth of top-$K$ recommendation lists.

\section{Experiment}
\begin{figure}[tb]
    \centering
    \begin{subfigure}[t]{0.4\textwidth}
        \centering
        \includegraphics[width=\linewidth]{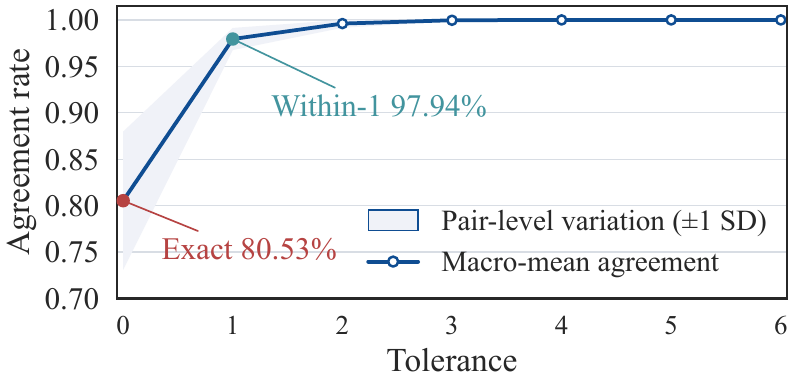}
        \caption{Agreement under score-difference tolerance.}
        \label{fig:tolerance_curve}
    \end{subfigure}
    \begin{subfigure}[t]{0.4\textwidth}
        \centering
        \includegraphics[width=\linewidth]{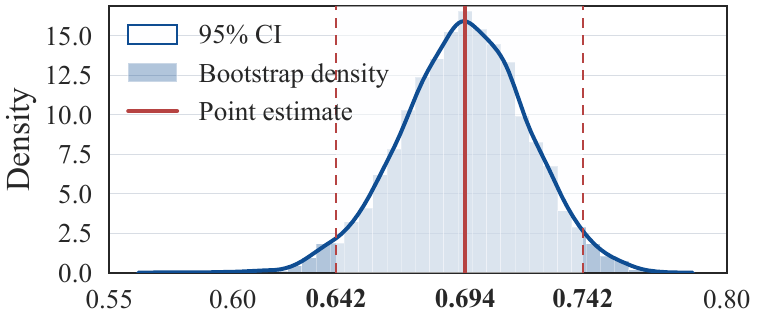}
        \caption{Bootstrap distribution of Ordinal Krippendorff's $\alpha$.}
        \label{fig:alpha_bootstrap}
    \end{subfigure}
    \caption{Human annotation reliability analysis, showing high agreement in their CDS annotations.}
    \label{fig:agreement}
\end{figure}

\subsection{Human Annotation Reliability}\label{sec:Human_Validation}
We assess the reliability of the independent human annotations collected during the construction of the gold set.
Specifically, we analyze the Step~\ding{172} annotations before conflict resolution, thereby evaluating whether different human evaluators can consistently apply the proposed CDS rubric without consensus-based adjustment, following prior evaluation practices~\cite{Pandalm,Video-bench}.
As shown in Figure~\ref{fig:agreement}, the human evaluators exhibit a high level of agreement in their CDS annotations, which remains strong even after accounting for chance agreement.
Overall, these results indicate that the proposed scoring rubric can be reliably applied by different human evaluators.

\begin{table}[tb]
\centering
\setlength{\tabcolsep}{0pt}
\renewcommand{\arraystretch}{0.65}

{\fontsize{8.9pt}{10pt}\selectfont
\begin{tabular*}{\linewidth}{
  l
  @{\hspace{0pt}}
  c
  @{\hspace{3pt}}
  c
  @{\hspace{3pt}}
  c
  @{\hspace{3pt}}
  c
  @{\hspace{3pt}}
  c
}
\toprule
{Model}
& \shortstack{{Spearman}$\uparrow$}
& \shortstack{{Kendall}$\uparrow$}
& \shortstack{{Pearson}$\uparrow$}
& \shortstack{{Exact}$\uparrow$}
& \shortstack{{MAE}$\downarrow$} \\
\midrule

\texttt{GLM-5.1}
& 0.6658
& 0.6410
& 0.7663
& 71.67\%
& 0.3337 \\

\texttt{GPT-5.5}
& 0.6414
& 0.6108
& 0.7312
& 67.35\%
& 0.4418 \\

\texttt{Gemini3.1-Pro}
& \textbf{0.7270}
& \textbf{0.7020}
& 0.7683
& 76.71\%
& 0.3097 \\

\texttt{Kimi-K2.6}
& 0.7003
& 0.6714
& 0.7610
& 71.43\%
& 0.3505 \\

\texttt{MiMo-V2.5-Pro}
& 0.6356
& 0.6027
& 0.6944
& 67.35\%
& 0.4538 \\

\texttt{Qwen3.7-Max}
& 0.7177
& 0.6960
& \textbf{0.7934}
& \textbf{78.03\%}
& \textbf{0.2605} \\

\bottomrule
\end{tabular*}
}

\caption{Comparison of LLM evaluators against human CDS judgments.
$\uparrow$ and $\downarrow$ indicate that higher and lower values are better, respectively, and \textbf{bold} denotes the best result.}
\label{tab:correlation_results}
\end{table}

\subsection{LLMs-Human Agreement} \label{sec:LLMs-Human Agreement}

To identify the most suitable LLMs for our evaluation protocol, we compare six leading models from the leaderboard\footnote{\url{https://artificialanalysis.ai/leaderboards/models}}: three open-weight models, \texttt{Kimi-K2.6} \cite{moonshot2026kimik26}, \texttt{MiMo-V2.5-Pro} \cite{xiaomi2026mimov25pro}, and \texttt{GLM-5.1} \cite{zai2026glm51}; and three proprietary models, \texttt{Gemini 3.1-Pro} \cite{google2026gemini31pro}, \texttt{GPT-5.5} \cite{openai2026gpt55}, and \texttt{Qwen3.7-Max} \cite{qwen2026qwen37max}.
Following prior work~\cite{G-eval,Prometheus,Flask}, we evaluate all models on the same gold set and compare their scores $s_i$ with human annotations using Spearman's $\rho$, Kendall's $\tau$, Pearson correlation, exact-match accuracy, and MAE.
As shown in Table~\ref{tab:correlation_results}, \texttt{Qwen3.7-Max} achieves the strongest human alignment in three of the five metrics and is therefore adopted as the default evaluator.
Most LLM models also correlate well with the human scores, suggesting that our protocol enables LLMs to reproduce judgments broadly shared by human evaluators.
Additional agreement results are reported in Appendix~F.

\subsection{Analysis of CDS} \label{sec:CDS_Analysis}
\begin{figure}[tb]
    \centering
    \includegraphics[width=0.9\linewidth]{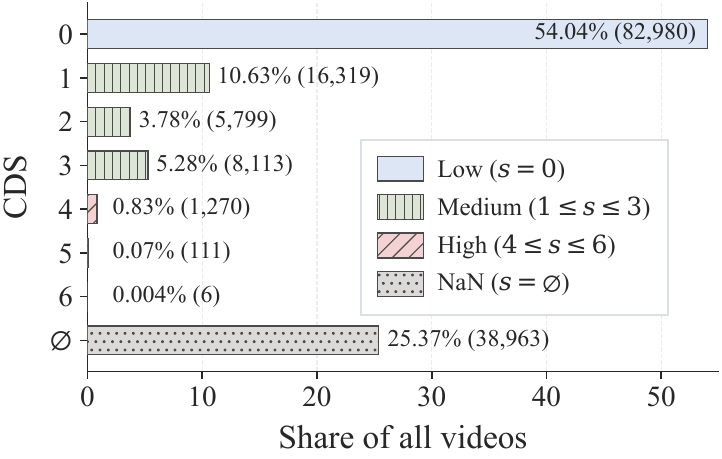}
    \caption{Distribution of CDS across all videos.
    Most videos fall into the low-CDS group, with few in the high-CDS group.}
    \label{fig:CDS_distribution}
\end{figure}
\textbf{CDS Distribution.}
Figure~\ref{fig:CDS_distribution} shows that the majority of videos fall into the low-CDS or NaN group\footnote{NaN cases mainly result from missing raw videos or ASR transcripts that are too short or noisy for reliable CDS assessment.}.
Overall, the distribution suggests that a large ratio of videos in short-video platforms imposes limited content depth.
This observation is consistent with the attention-economy nature of platforms, where entertaining and attention-grabbing content is prevalent.

\begin{table}[t]
    \centering
    \small
    \setlength{\tabcolsep}{3.5pt}
    \renewcommand{\arraystretch}{0.7}
    \begin{tabular}{clll}
        \toprule
        \textbf{Rank}
        & \textbf{Low}
        & \textbf{Medium}
        & \textbf{High} \\
        \midrule
        1 & Dance          & Health           & Finance     \\
        2 & Comedy         & Law              & Military    \\
        3 & Beauty         & Science          & History     \\
        4 & Music          & History          & Law         \\
        5 & Short Dramas   & Finance          & Science     \\
        6 & Casual Videos  & Real Estate      & Information \\
        7 & Anime          & Digital Products & Health      \\
        \bottomrule
    \end{tabular}
    \caption{Top seven categories in each CDS group, ranked by proportions.
    Low-CDS categories differ from the others, while medium- and high-CDS categories largely overlap.}
    \label{tab:cds_top7}
\end{table}

\textbf{Category Distribution.}
Table~\ref{tab:cds_top7} presents the top categories across three CDS group. We observe that medium- and high-CDS videos are more frequently associated with knowledge-intensive categories, such as Finance, History, and Law.
In contrast, low-CDS videos are more commonly associated with entertainment-oriented categories, such as Dance, Comedy, and Beauty.
These category-level patterns are consistent with our design intuition of CDS: videos involving domain knowledge tend to receive higher CDS, whereas videos primarily designed for entertainment tend to receive lower CDS.
See Appendix~G for distribution details.

\begin{figure}[tb]
    \centering
    \includegraphics[width=1\linewidth]{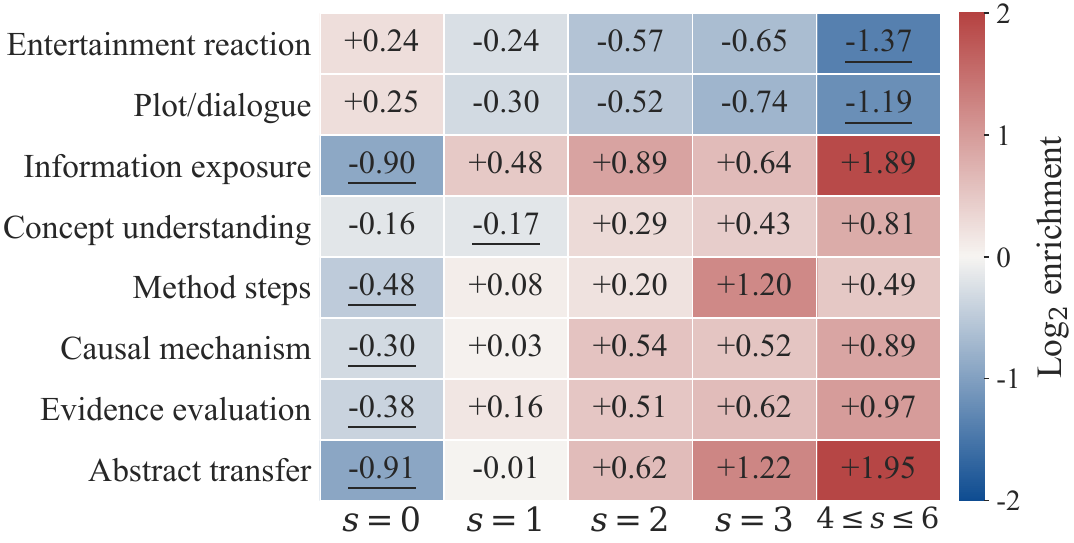}
    \caption{Lexicon-based enrichment across CDS levels.
    Each cell reports the relative enrichment of signal words at CDS levels, with positive and negative values indicating values above and below the overall average, respectively.
    The smallest enrichment value in each row is shown in \underline{underlined}.}
    \label{fig:singal_heatmap}
\end{figure}

\textbf{CDS Word Frequency.}
Figure~\ref{fig:singal_heatmap} shows the enrichment patterns of different CDS levels based on our predefined theory-oriented lexicon\footnote{We built the lexicon by extracting the top 500 words in gold subset and the top 200 words at each CDS level, grouping them into theoretical categories, and expanding each category.}.
Entertainment reaction and plot/dialogue show a decreasing trend as the CDS level rises, whereas most other categories exhibit the opposite trend.
These findings are consistent with our expectations: low-CDS videos are more likely to focus on entertainment-oriented reactions and plot-level descriptions, whereas high-CDS videos are more likely to contain signals related to explanation, evidence evaluation, generalization, transfer, and conceptualized expression.
We also provide details representative words for each CDS level in Appendix~H.

\begin{table}[t]
\centering
\begin{threeparttable}
\setlength{\tabcolsep}{3pt}
\renewcommand{\arraystretch}{0.75}
{\fontsize{9pt}{10pt}\selectfont
\begin{tabular*}{\linewidth}{
  @{\extracolsep{\fill}}
  l
  c
  c
  @{}
}
\toprule
\textbf{Variable}
& \textbf{Association}
& \textbf{Explained variance} \\
\midrule

Caption length
& Spearman $\rho = 0.0828^{*}$
& $R^2 = 0.57\%$ \\

Log-ASR length\tnote{a}
& Spearman $\rho = 0.2182^{*}$
& $R^2 = 6.09\%$ \\

Category
&Cramér $V = 0.2469^{*}$
& $\eta^2 = 25.34\%$ \\

\bottomrule
\end{tabular*}
}

\begin{tablenotes}[flushleft]
{\fontsize{8pt}{10pt}\selectfont
\item[a]
ASR length $|a|$ exhibits a long-tailed distribution.
Therefore, we use log-transformed form
$\log(1+|a|)$ to estimate the relationship.
\item $^{*}$ denotes statistical significance at $p < 0.001$.
}
\end{tablenotes}
\end{threeparttable}
\caption{
Associations between valid CDS ($s \neq \varnothing$) and surface-level input attributes.
CDS is more strongly associated with ASR length and category than with caption length.
}
\label{tab:cpd_surface_attributes}
\end{table}

\textbf{Surface-Level Correlates of CDS.}
Table~\ref{tab:cpd_surface_attributes} summarizes the relationships between valid CDS ($s\neq \varnothing$) and three surface-level input attributes: Caption length, ASR length, and category.
The results show that caption length has only a weak association with CDS, whereas ASR length and category exhibit stronger associations.
This pattern is consistent with high-CDS content requiring sufficient textual space to express explanations, procedures, and reasoning structures.
Category differences may similarly arise from their inherent content orientation, with entertainment-oriented and knowledge-oriented categories tending to receive lower and higher CDS scores, respectively, as shown in Table~\ref{tab:cds_top7}.

\subsection{Evaluation Protocol Robustness}
In Section~\ref{sec:CDS_Analysis}, we observe that CDS exhibit certain correlations with video category and ASR length.
These observations naturally raise two robustness concerns: \textit{Whether the evaluation protocol has category prior bias and verbosity bias}.
To examine these issues, we conduct paired counterfactual robustness tests \cite{LLM-as-a-judge}, where only one input field is perturbed at a time.
For each video $i$, we compare the original CDS score $s_i$ and ASR length $|a_i|$ with their counterfactual values $s_i'$ and $|a_i'|$. We define the absolute score change and relative ASR length change as
$\Delta s_i = |s_i' - s_i|$ and $ \delta |a_i| = {(|a_i'| - |a_i|)}/{|a_i|}$, respectively.

\begin{table}[t]
\centering
\renewcommand{\arraystretch}{0.65}
\setlength{\tabcolsep}{1.5pt}
{\fontsize{9pt}{10pt}\selectfont
\begin{tabular*}{\linewidth}{
  @{\extracolsep{\fill}}
  l c c c c
  @{}
}
\toprule
\textbf{Perturbation}
& \textbf{$N$}
& \textbf{P($\Delta s = 1$)}
& \textbf{P($\Delta s = 2$)}
& \textbf{P($\Delta s > 2$)}
\\
\midrule
Low $\rightarrow$ Medium
& 50 & 0.00\% & 0.00\% & 0.00\%  \\

Low $\rightarrow$ High
& 50 & 0.00\% & 0.00\% & 0.00\%  \\

\midrule
Medium $\rightarrow$ Low
& 60 & 6.67\% & 3.33\% & 0.00\% \\

High $\rightarrow$ Low
& 46 & 13.04\% & 13.04\% & 0.00\% \\
\bottomrule
\end{tabular*}
}
\caption{
Category counterfactual perturbation results measured by CDS shift
magnitude.
$\text{Source}\rightarrow\text{Target}$ indicates perturbing videos
from the Source CDS group to Target-associated categories.
Category perturbations leave low-CDS videos unchanged and cause only limited one- or two-level shifts for medium- and high-CDS videos.
}
\label{tab:category_shift_probability}
\end{table}

\textbf{Robustness to Category Prior Bias.}
We examine whether the evaluation protocol relies on category-level shortcuts by overemphasizing category labels while underutilizing ASR transcripts, which provide more direct evidence of a video's semantic content.
As shown in Table~\ref{tab:category_shift_probability},
changing low-CDS videos to either medium- or high-CDS-associated categories does not change their CDS.
For the reverse direction, mapping high- or medium-CDS videos to low-CDS-associated categories may introduce slight downward shifts, with more pronounced changes observed for high-CDS videos.
This is consistent with our conservative scoring principle: when the available evidence does not fully support a higher CDS level, the evaluator tends to assign a lower score.

\begin{table}[t]
\centering
\renewcommand{\arraystretch}{0.65}

{\fontsize{8.5pt}{10pt}\selectfont
\setlength{\tabcolsep}{0pt}

\begin{tabular}{
  c
  @{\hspace{5pt}}
  c
  @{\hspace{5pt}}
  c
  @{\hspace{5pt}}
  c
  @{\hspace{1pt}}
  c
  @{\hspace{1pt}}
  c
  @{\hspace{1pt}}
  c
}
\toprule
\textbf{Perturbation}
& \textbf{Group}
& \textbf{$N$}
& \textbf{$\delta |a|$}
& \textbf{P($\Delta s = 1$)}
& \textbf{P($\Delta s = 2$)}
& \textbf{P($\Delta s > 2$)} \\
\midrule

\multirow{2}{*}{ASR length $\uparrow$}
& Low
& 50
& +52.05\%
& 2.00\%
& 0.00\%
& 0.00\% \\

& Medium
& 60
& +51.39\%
& 5.00\%
& 0.00\%
& 0.00\% \\

\midrule

\multirow{2}{*}{ASR length $\downarrow$}
& Medium
& 60
& -24.49\%
& 8.33\%
& 0.00\%
& 0.00\% \\

& High
& 46
& -27.35\%
& 23.91\%
& 15.22\%
& 0.00\% \\

\bottomrule
\end{tabular}
}

\caption{
ASR counterfactual perturbation results measured by CDS shifts.
$\Delta s_i$ and $\delta |a_i|$ denote the absolute CDS change and relative ASR length change, respectively.
CDS remains largely stable under lengthening and changes modestly under shortening, with no shifts exceeding two levels.
}
\label{tab:asr_shift_probability}
\end{table}

\textbf{Robustness to Verbosity Bias.}
We examine whether the evaluation protocol favors longer ASR transcripts without additional meaningful information.
Table~\ref{tab:asr_shift_probability} shows that meaningless length expansion has minimal impact, whereas transcript shortening affects high-CDS videos more than medium-CDS videos, with no CDS change exceeding two levels.
This sensitivity is consistent with our conservative scoring principle, as transcript compression may weaken the reasoning evidence required for high CDS levels, including argument logic and evidence connections.

Overall, these results suggest that our evaluation protocol primarily responds to meaningful informational and reasoning structures in the videos, rather than superficial cues such as category labels or ASR transcript length.
Details and stability experiment are provided in Appendix~I.

\section{SCOPE-Bench Leaderboard}
\begin{table}[t]
\centering
\footnotesize
\setlength{\tabcolsep}{1pt}
\renewcommand{\arraystretch}{0.65}
{\fontsize{9pt}{10pt}\selectfont
\begin{tabular*}{\linewidth}{
  @{\extracolsep{\fill}}
  l
  *{6}{c}
  @{}
}
\toprule
\multirow{2}{*}{\textbf{Method}}
& \multicolumn{3}{c}{\textsc{ShortVideoSampled}}
& \multicolumn{3}{c}{\textsc{ShortVideoFull}} \\
\cmidrule(lr){2-4}
\cmidrule(lr){5-7}

& R@20$\uparrow$
& A@20$\uparrow$
& E@20$\uparrow$
& R@20$\uparrow$
& A@20$\uparrow$
& E@20$\uparrow$ \\

\midrule

\multicolumn{7}{l}{\textit{\textbf{ID-based Recommendation}}} \\

BPR
& 3.30
& 6.90
& 6.85
& 2.22
& 7.06
& 7.10 \\

NCF
& 3.01
& 6.49
& 6.42
& 2.09
& 7.46
& 7.53 \\

LightGCN
& 3.54
& 7.36
& 7.33
& 2.37
& 6.89
& 7.08 \\

\midrule

\multicolumn{7}{l}{\textit{\textbf{Multimodal Recommendation}}} \\

VBPR
& 2.73
& 6.22
& 5.70
& 1.80
& 7.44
& 7.95 \\

GRCN
& 2.83
& 7.92
& 7.85
& 1.64
& 6.72
& 6.73 \\

LATTICE
& 2.70
& 7.13
& 7.04
& 2.25
& 6.97
& 7.10 \\

BM3
& 2.88
& 6.72
& 6.66
& 2.41
& 6.70
& 6.80 \\

FREEDOM
& 3.30
& 7.13
& 7.30
& 2.51
& 6.81
& 6.84 \\

MGCN
& \textbf{3.82}
& 7.38
& 7.58
& \textbf{2.65}
& 6.89
& 6.99 \\

LGMRec
& 3.40
& 6.71
& 6.63
& 2.42
& 7.24
& 7.52 \\

DiffMM
& 3.24
& 7.24
& 7.35
& 2.39
& 7.10
& 7.18 \\

REARM
& 3.76
& 7.71
& 7.76
& 2.42
& 6.81
& 6.93 \\

FITMM
& 3.25
& \textbf{8.12}
& \textbf{8.35}
& 2.41
& \textbf{7.59}
& \textbf{7.88} \\

\midrule

Random
& 0.08
& 6.90
& 6.90
& 0.02
& 6.59
& 6.59 \\

\bottomrule
\end{tabular*}
}

\caption{
Engagement and content-depth performance of baselines on two datasets. All metrics are scaled by $\times 100$, where R denotes Recall, A/E denote A-LCDS/E-LCDS, and the best results are \textbf{bolded}. Baselines are competitive on engagement but remain in the low-LCDS range and close to random recommendation on content-depth metric.
}
\label{tab:baseline_results}
\end{table}

We evaluate 13 baselines on SCOPE-Bench, including three ID-based methods,
BPR \cite{BPR}, NCF \cite{NCF}, and LightGCN \cite{LightGCN},
and ten multimodal methods, VBPR \cite{VBPR}, GRCN \cite{GRCN}, LATTICE \cite{LATTICE}, BM3 \cite{BM3}, FREEDOM \cite{FREEDOM}, MGCN \cite{MGCN}, LGMRec \cite{LGMRec}, DiffMM \cite{DiffMM}, REARM \cite{REARM}, and FITMM \cite{FITMM}.
The Random baseline reports the expected performance of uniformly sampling items from each candidate set.
We use an 8:1:1 training-validation-test split \cite{ShortVideo}.
As shown in Table~\ref{tab:baseline_results}, existing methods achieve competitive performance on engagement metrics, whereas their A-LCDS and E-LCDS scores consistently remain within the Low-LCDS range, i.e, $[0,1/6)$.
Moreover, most methods perform close to the Random baseline.
These results indicate that stronger engagement performance does not necessarily translate into the recommendation of content with greater depth.
Appendix~C presents the experimental setup, results under an alternative treatment of $s_i=\varnothing$, training trajectories of engagement and content-depth metrics, and CDS-aware optimization.
These results further demonstrate that engagement and content depth are currently decoupled.

\section{Conclusion}
In this paper, we propose a new metric, termed \textit{CDS}, to measure the content depth of short videos.
CDS provides a principled basis for evaluating and optimizing content depth in existing RSs.
To comprehensively evaluate this dimension, we construct SCOPE-Bench, the first benchmark that supports both item- and list-level content-depth evaluation.
Empirical results demonstrate the interpretability, practical utility, and robustness of the proposed evaluation framework.
Our experiments further reveal that existing RSs tend to favor low-depth videos, and remain limited in recommending videos with high CDS.
Hereby, CDS and SCOPE-Bench establish a new evaluation axis for developing short-video RSs that jointly consider content depth and user engagement.

\bibliography{reference}

\end{document}